\pdfoutput=1

\documentclass[11pt,a4paper]{article}

\usepackage[hyperref]{EMNLP2022}

\usepackage{times}
\usepackage{latexsym}

\usepackage[T1]{fontenc}

\usepackage[utf8]{inputenc}

\usepackage{microtype}

\usepackage{inconsolata}

\usepackage{graphicx}
\usepackage{xcolor}
\usepackage{comment}
\usepackage{booktabs}
\usepackage{amsmath}
\usepackage{amssymb}
\usepackage{amsthm}
\usepackage{bbm} 
\usepackage{subcaption}
\usepackage{multirow}
\usepackage{comment}
\usepackage{makecell} 
\usepackage{tabularx}
\usepackage{CJKutf8} 
\usepackage[normalem]{ulem}
\useunder{\uline}{\ul}{}

\usepackage{forest}
\definecolor{folderbg}{RGB}{124,166,198}
\definecolor{folderborder}{RGB}{110,144,169}
\def\Size{4pt}
\tikzset{
  folder/.pic={
    \filldraw[draw=folderborder,top color=folderbg!50,bottom color=folderbg]
      (-1.05*\Size,0.2\Size+5pt) rectangle ++(.75*\Size,-0.2\Size-5pt);  
    \filldraw[draw=folderborder,top color=folderbg!50,bottom color=folderbg]
      (-1.15*\Size,-\Size) rectangle (1.15*\Size,\Size);
  }
}

\usepackage[export]{adjustbox}
\usepackage{paralist}

\title{Fengshenbang 1.0: Being the Foundation of Chinese Cognitive Intelligence}

\author{
Project Manager: 
{\bf Jiaxing Zhang\thanks{\ \ Corresponded author.}, }
{\bf Ruyi Gan\footnotemark[1]} \\
{\bf Junjie Wang, }
{\bf Yuxiang Zhang, }
{\bf Lin Zhang, }
{\bf Ping Yang, }
{\bf Xinyu Gao, }
{\bf Ziwei Wu, }\\
{\bf Xiaoqun Dong, }
{\bf Junqing He, }
{\bf Jianheng Zhuo, }
{\bf Qi Yang, }
{\bf Yongfeng Huang,}\\
{\bf Xiayu Li, }
{\bf Yanghan Wu, }
{\bf Junyu Lu, }
{\bf Xinyu Zhu, }
{\bf Weifeng Chen, }
{\bf Ting Han,}\\
{\bf Kunhao Pan, }
{\bf Rui Wang, }
{\bf Hao Wang, }
{\bf Xiaojun Wu, }
{\bf Zhongshen Zeng,}
{\bf Chongpei Chen }\\
CCNL, IDEA, Shenzhen, China\\
\url{https://github.com/IDEA-CCNL/Fengshenbang-LM}
}

\begin{document}
\begin{CJK*}{UTF8}{gbsn}

\maketitle


\begin{abstract}
Nowadays, foundation models become one of fundamental infrastructures in artificial intelligence, paving ways to the general intelligence.
However, the reality presents two urgent challenges: existing foundation models are dominated by the English-language community; users are often given limited resources and thus cannot always use foundation models.
To support the development of the Chinese-language community, we introduce an open-source project, called Fengshenbang, which leads by the research center for Cognitive Computing and Natural Language (CCNL).
Our project has comprehensive capabilities, including large pre-trained models, user-friendly APIs, benchmarks, datasets, and others.
We wrap all these in three sub-projects: the Fengshenbang Model, the Fengshen Framework,  and the Fengshen Benchmark.

An open-source roadmap, Fengshenbang, aims to re-evaluate the open-source community of Chinese pre-trained large-scale models, prompting the development of the entire Chinese large-scale model community.
We also want to build a user-centered open-source ecosystem to allow individuals to access the desired models to match their computing resources.
Furthermore, we invite companies, colleges, and research institutions to collaborate with us to build the large-scale open-source model-based ecosystem.
We hope that this project will be the foundation of Chinese cognitive intelligence.

\textit{Note that this report also has a Chinese-language version (Starting from Section~\ref{sec:intro_zh}).}

\end{abstract}






\section{Introduction}
\label{sec:intro}

Remarkable advances in Artificial Intelligence (AI) have produced great models, in particular, pre-training based foundation models~\cite{DBLP:journals/corr/abs-2108-07258/opportunities} become an emerging paradigm. 
In contrast to traditional AI models that must be trained on vast datasets for one or a few scenarios, foundation models can be adapted to a wide range of downstream tasks, therefore, limiting the amount of resource demanded to acquire an AI venture off the ground. 
Moreover, we observe that these models grow rapidly within a short period, around $10$ times each year. 
For instance, BERT~\cite{DBLP:conf/naacl/DevlinCLT19/bert} has $100$ million parameters and GTP-3~\cite{DBLP:conf/nips/BrownMRSKDNSSAA20/gpt3} has over $100$ billion parameters. 
Many of the forefront challenges in AI, especially generalization ability, are becoming achievable due to this inspiring trend.

Foundation models, most notably language models, are dominated by the English-language community. 
The Chinese language as the world's largest spoken language (native speakers), however, has no systematic research resources to support it, making the progress in the Chinese language domain lag behind others. 
To address this urgent need, we develop a Chinese language driven foundation ecosystem, named Fengshenbang, that incorporates pre-trained models, task-specific fine-tune applications, benchmarks, and datasets.

\begin{figure}[t]
  \centering
  \includegraphics[width=0.5\textwidth]{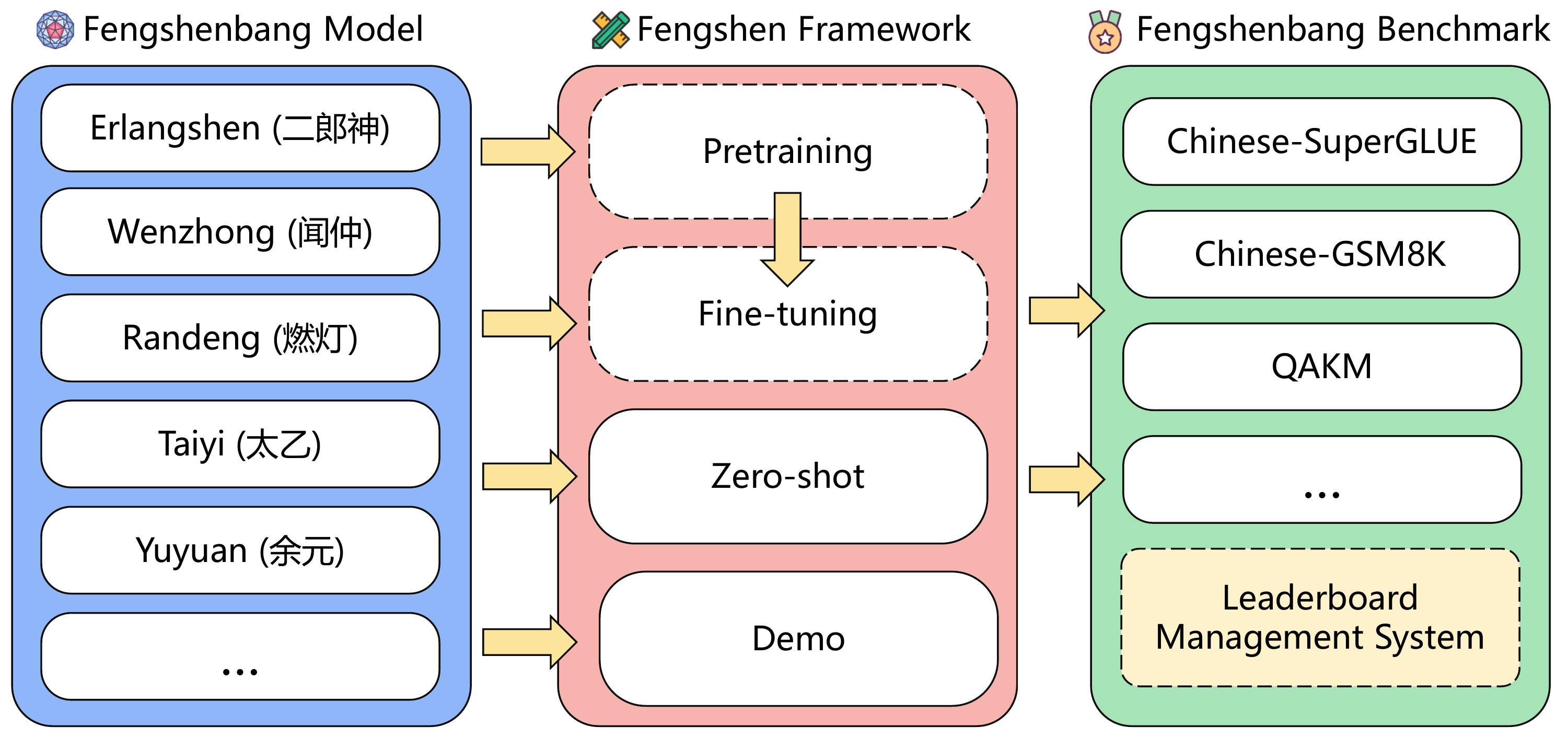}
  \caption{The overview of Fengshenbang project.}
  \label{fig:introduction}
\end{figure}

Our goal is to build a comprehensive, standardized and user-centered ecosystem. 
Although this can be instantiated in a variety of ways, we present the following design that we find to be particularly effective.
Concretely, as illustrated in Figure~\ref{fig:introduction}, Fengshenbang ecosystem has three core modules as follows.
\begin{itemize}
    \item Fengshenbang Model (Section~\ref{sec:model})
    \item Fengshen Framework (Section~\ref{sec:framework})
    \item Fengshenbang Benchmark (Section~\ref{sec:benchmark})
\end{itemize}

Before discussing the Fengshenbang models, we first lay out some preliminaries.
The NLP community has broad research interests, which can be categorized into two main types: general-purpose tasks and special-purpose tasks.
The former includes Natural Language Understanding (NLU), Natural Language Generation (NLG), and Natural Language Transformation (NLT); the latter covers multimodality, specific domains, and others.
Importantly, we argue that a platform with comprehensive foundation models is necessary, therefore, we consider all of these in Fengshenbang models.
Further, we provide associated models to fine-tune for downstream tasks, so that users with limited resources can access foundation models without effort.

Next, we introduce the Fengshen framework to be user-centered, allowing users to further refine or modify models w.r.t their purposes based on our resources.
More specifically, we provide a flexible menu in order to enable low-cost utilization, including standardized data processors, detailed tutorial examples, a docker-alike environment, and industry-standard APIs.

Last but not the least, our proposed ecosystem includes a benchmark module, allowing users to perform fair comparisons and the whole community to track progress.
In particular, we choose to open source the leaderboard system in the future to make the comparison fair and promote the development of more customized Leaderboard system. 

To this end, we have introduced our ecosystem.
Although this seems complicated, with only $3$ sequential steps, users can build their applications based on our resources.
\begin{enumerate}
    \item[Step 1:] Choosing a pre-trained Chinese NLP model from our open-source library of Fengshenbang Models.
    \item[Step 2:] Employing Fengshen Framework to adjust the model by exploring the our tutorial examples.
    \item[Step 3:] Evaluating on downstream tasks, such as Fengshenbang Benchmarks or custom tasks.
\end{enumerate}

\section{Fengshenbang Model}
\label{sec:model}

In this section, we will introduce the details of building Fengshenbang Models. 
As shown in Table~\ref{table:user_tax}, we present a User-Centered Taxonomy (UCT) (Section~\ref{sec:model_taxonomy}), aiming to classify the requirement of users and designing corresponding models. 
More details of these models are included in Section~\ref{sec:model_selelction}, which explains our model selection criteria to ensure the quality.
In fact, we provide a nomenclature in Section~\ref{sec:naming_convention} to reduce the misunderstanding as low as possible, where users can locate the desired models based on the nomenclature. 
More details of these models can be find from Section~\ref{sec:erlangshen} to Section~\ref{sec:tbd}. 
Please refer to Appendix~\ref{append:all_model} for more details of $49$ models.

\subsection{Model Design}


\begin{table}[]
\small
\centering
\begin{tabular}{l|c|c|c}
\toprule
Demand                   & Task        & Series     & Example Model \\ \midrule
\multirow{3}{*}{General} & NLU         & Erlangshen & BERT, DeBERTa          \\
                         & NLG         & Wenzhong   & GPT2          \\
                         & NLT         & Randeng    & T5, BART            \\ \midrule
\multirow{3}{*}{Special} & MM          & Taiyi      & CLIP          \\
                         & Domain      & Yuyuan     & BioBERT    \\
                         & Exploration & TBD  & TBD      \\ \bottomrule
\end{tabular}
\caption{User-centered taxonomy with example models. ``TBD'' indicates ``To Be Discussed''.}
\label{table:user_tax}
\end{table}

\subsubsection{User-centered Taxonomy (UCT)}
\label{sec:model_taxonomy}

Advances in NLP have developed numerous powerful models from different perspectives, including research and applications. 
To understand the difference and track progress, there is an opportunity to standardize taxonomy in this field. 
However, models are often difficult to be categorized due to their complexity.
For example, TinyBERT~\cite{DBLP:conf/emnlp/JiaoYSJCL0L20/tinybert} can be classified as ``Encoder-only'' in the model architecture, or it can be assigned as ``Distillation'' under model parameter reduction methods. 
To reduce misunderstanding, we introduce the User-centered Taxonomy (UCT), which consults numerous NLPers. 
In general demands, there are common NLP tasks, which are classified into Natural Language Understanding (NLU), Natural Language Generation (NLG), and Natural Language Transformation (NLT). 
Due to the fast development, NLP community brings special demands to the entire AI community, which are often assigned to MultiModal (MM), Domains and Exploration. 
Moreover, we assign a series name for each task. 
Note that we will update UCT timely according to the development of the NLP field.

\noindent\textbf{Natural Language Understanding (NLU)}

NLU tasks make use of syntactic and semantic analysis of text to understand the meaning of sentences. 
The syntax is related to the grammatical structure of a sentence, and semantics refers to its intended meaning. 
Relationships between words and phrases are also important as these will lead to different concepts.
In addition, some problems in this task are difficult to solve even for humans. 
To evaluate the performance of NLU, several tasks are developed to ensure reliability:

\begin{itemize}
    \item Semantic Matching
    \item Sentiment Analysis
    \item Natural Language Inference
    \item Entity Recognition
    \item Relationship Extraction
    \item Event Extraction
    \item Chinese Word Segmentation
    \item \dots
\end{itemize}

\noindent\textbf{Natural Language Generation (NLG)}

Different from computer reading comprehension in NLU, NLG is concerned with developing computer systems that produce understandable writing.
An NLG-capable system should be able to generate natural language by forming its ideas as opposed to transforming existing data.
And, the generated text needs to be coherent and understandable to humans.
We assign several NLG tasks as follows.
\begin{itemize}
    \item Creative Writing
    \item Causal Reasoning
    \item Controlled Generation
    \item Multi-Step Reasoning
    \item \dots
\end{itemize}

\noindent\textbf{Natural Language Transformation (NLT)}

We define NLT tasks as source-to-target transformation tasks.
In contrast to NLG, NLT is based on the source objects and target objects.
Language models require generating or transforming target objects by understanding source objects.
Taking machine translation as an example, given a text in one language, an AI system needs to generate the corresponding text in another language.
In summary, we list the NLT tasks as follows.

\begin{itemize}
    \item Machine Translation
    \item Text Summarization
    \item Text Simplification
    \item Grammatical Error Correction
    \item Question Answering
    \item Dialogue System
    \item \dots
\end{itemize}

\noindent\textbf{MultiModal (MM)}

Due to the growing demand for complex scenarios, unimodal models cannot handle multiple modalities.
Transformer-based Pre-trained Language Models (PLMs) are widely used in fields like computer vision and audio processing due to their flexible architecture.
In addition, cognitive intelligence requires intelligent systems to learn from multiple modalities, including text, images, and audio.
To this end, several multimodal scenarios are introduced,
such as text-to-image generation and multimodal semantic understanding.

\begin{itemize}
    \item Text-to-image Generation
    \item Image Captioning
    \item Cross-modal Retrieval
    \item Visual Question Answering
    \item Automatic Speech Recognition
    \item Text-to-speech
    \item Voice Conversation
    \item Protein Structure Prediction
    \item \dots
\end{itemize}

\noindent\textbf{Domain}

Notably, PLMs have achieved phenomenal success in a variety of specific domains.
Continuous pre-training is a key advantage of constructing domain-specific models.
Because the model is not trained from start, it results in less computational resource consumption.
Some domains and several related models are listed below:

\begin{itemize}
    \item Finance: FinBERT~\cite{DBLP:journals/corr/abs-2006-08097/finbert}
    \item Biomedical: BioBERT~\cite{DBLP:journals/bioinformatics/LeeYKKKSK20/biobert}, ClinicalBERT~\cite{DBLP:journals/corr/abs-1904-03323/clinical_bert}, PubMedBERT~\cite{DBLP:journals/health/GuTCLULNGP22/pubmed_bert}
    \item Legal: LEGAL-BERT~\cite{DBLP:journals/corr/abs-2010-02559/legalbert}, ALeaseBERT~\cite{DBLP:journals/corr/abs-2010-10386/aleasebert}
    \item Programming: CoTexT~\cite{DBLP:journals/corr/abs-2105-08645/cotext}, CodeBERT~\cite{DBLP:conf/emnlp/FengGTDFGS0LJZ20/codebert}, GraphCodeBERT~\cite{DBLP:conf/iclr/GuoRLFT0ZDSFTDC21/graphcodebert}, CodeGPT-adapted~\cite{DBLP:conf/nips/LuGRHSBCDJTLZSZ21/codegpt-adap}, Codex~\cite{DBLP:journals/corr/abs-2107-03374/codex}
    \item Academic: OAG-BERT~\cite{DBLP:journals/corr/abs-2103-02410/oagbert}, MathBERT~\cite{DBLP:journals/corr/abs-2105-00377/mathbert}, SciBERT~\cite{DBLP:conf/emnlp/BeltagyLC19/scibert}
    \item \dots
\end{itemize}

\noindent\textbf{Exploration}

Together with other organizations, such as technology companies and universities, we will develop some experimental models in NLP.


\subsubsection{Model Selection}
\label{sec:model_selelction}

Since many papers and models are proposed every year, we select only a few of them for pre-training and then open source, aiming to control the overall quality and use computing resources wisely.
We select models based on the following rules:

\noindent\textbf{Powerful.} 
Some models present astonishing performance on downstream tasks.
These are either published in English version or even not released.
In addition, they are often not matching with for Chinese-language community, but need be extended or modified.

\noindent\textbf{Diversity.} 
Multiple models are widely used in various NLP tasks, such as BERT~\cite{DBLP:conf/naacl/DevlinCLT19/bert}, GPT-3~\cite{DBLP:conf/nips/BrownMRSKDNSSAA20/gpt3} and Transformer~\cite{DBLP:conf/nips/VaswaniSPUJGKP17/transformer}. 
Moreover, they are often to be extended and adapted easily.
We include these models when considering diverse scenarios such as different downstream tasks, architectures, model sizes and pre-training methods.

\noindent\textbf{Usability.} Open-sourced models should easy to be understood and implemented in practice.
Additionally, users can use some models out-of-the-box for desired downstream tasks.


\subsubsection{Naming Convention}
\label{sec:naming_convention}

To better understand Fengshenbang models, we introduce the naming namespace with the following template:
\begin{equation}
\small
Name \in \{ Series-Model-Parameter-Extra \}^N
\end{equation}
Here, $Series$ is the name of series that come from a list of gods in a Chinese fiction called Fengshenbang~\cite{hsun2000brief/fengshenbang}. 
Each series manages a category of NLP task. 
$Model$ represents the structure of models, such as BERT and GPT-3. 
$Parameter$ is the number of parameters and $N$ is the total number of our open-sourced models. 
$Extra$ indicates extra information of settings, such as fine-tuning on downstream datasets. 
For example, ``Erlangshen-Roberta-110M-NLI'' shows this is in Erlangshen series to solve NLU tasks. 
The model structure is based on RoBERTa with 110M parameters. 
After pre-training on Chinese datasets, we fine-tune it on NLI tasks.


\subsection{Erlangshen (NLU)}
\label{sec:erlangshen}

The Erlangshen series is designed to solve NLU problems, including Megatron-BERT, ZEN, RoBERTa, DeBERTa, Longformer, UBERT and UnifiedMC.

\subsubsection{MegatronBERT}

For training a billion-sized BERT, we follow the instructions of Megatron-LM~\cite{DBLP:journals/corr/abs-1909-08053/Megatronlm} and pre-train the BERT~\cite{DBLP:conf/naacl/DevlinCLT19/bert} on WuDao Corpora (180 GB version)~\cite{DBLP:journals/aiopen/YuanZDDLCZYT21/wudao_corpora}. 
Given the Chinese grammatical structure and the difficulty of training in large-scale model, we apply the following four pre-training strategies to improve BERT.\\
(1) Whole Word Masking (WWM). 
We adopt WWM~\cite{DBLP:journals/taslp/CuiCLQY21/wwm} by considering the linguistic features of Chinese, which is to process whole Chinese words instead of individual Chinese characters in popular WordPiece tokenizer.\\
(2) Knowledge-based Dynamic Masking (KDM). 
Instead of random masking in Masked Language Modeling (MLM), we attempt to mask token-rich semantic information to yield effective and efficient models.\\
(3) Sentence Order Prediction (SOP). 
Referring to ALBERT~\cite{DBLP:conf/iclr/LanCGGSS20/albert}, the LMs learn the inter-sentence information to get powerful representations. 
Hence, as a pre-training task, Erlangshen-MegatronBert models employ SOP instead of NSP.\\
(4) Pre-layer Normalization (Pre-LN). 
After applying post-layer normalization in the LM pre-training phase, we find that the loss rises abnormally fast as the model size increases.
Therefore, we apply Pre-LN~\cite{DBLP:conf/icml/XiongYHZZXZLWL20/pre-ln} to overcome this issue.

To the best of our knowledge, the largest BERT in the Chinese open source community is Erlangshen-MegatronBert-1.3B when publicly released.
The Erlangshen-MegatronBert model has benefited many developers, as demonstrated by the 4.5K\footnotemark[1] monthly downloads of our Erlangshen-MegatronBert model.
Thanks to the excellent performance, Erlangshen-MegatronBert models gain three important mentions:\\
(1) On November 10, 2021, it topped the FewCLUE few-shot learning tasks on CLUE benchmark~\cite{DBLP:conf/coling/XuHZLCLXSYYTDLS20/clue}. 
Among them, our model outperformed human performance in CHID (idiom fill-in-the-blank) and TNEWS (news classification) subtasks.
In addition, it ranked the top in CHID (idiom fill-in-the-blank), CSLDCP (subject literature classification), and OCNLI (natural language inference) tasks.\\
(2) On January 24, 2022, it topped the ZeroCLUE zero-shot Learning task on CLUE benchmark. 
For each of these tasks, we ranked the first places in CSLDCP (Subject Literature Classification), TNEWS (News Classification), IFLYTEK (Application Description Classification), CSL (Abstract Keyword Recognition), and CLUEWSC (Coreference Resolution) tasks.\\
(3) We topped the CLUE benchmark semantic matching task on July 10, 2022, 2022~\cite{DBLP:journals/corr/abs-2208-02959/no1_simclue}.

\footnotetext[1]{The data was obtained on August 18, 2022}

\begin{itemize}
    \item Erlangshen-MegatronBert-1.3B
    \item Erlangshen-MegatronBert-1.3B-NLI
    \item Erlangshen-MegatronBert-1.3B-Sentiment
    \item Erlangshen-MegatronBert-1.3B-Similarity
    \item Erlangshen-MegatronBert-3.9B-Chinese
\end{itemize}

Next, we will make the paper publicly available for details on Erlangshen-MegatronBert.

\subsubsection{ZEN}

We open source and publicly release ZEN1~\cite{DBLP:conf/emnlp/DiaoBSZW20/zen1} and ZEN2~\cite{DBLP:journals/corr/abs-2105-01279/zen2} using our Fengshen Framework in collaboration with the team ZEN.
More precisely, by bringing together knowledge extracted by unsupervised learning, ZEN1 learns different textual granularity information through N-gram methods.
ZEN1 can obtain good performance gains by training only on a single small corpus (low-resource scenarios).
ZEN2 pre-trains the n-gram-enhanced encoders with large-scale datasets and special pre-training strategies.
In the next step, we continue with the ZEN team to explore the optimization of PLM and improve the performance on downstream tasks.

\begin{itemize}
    \item Erlangshen-ZEN1-224M-Chinese
    \item Erlangshen-ZEN2-345M-Chinese
    \item Erlangshen-ZEN2-668M-Chinese
\end{itemize}

\subsubsection{RoBERTa}

To obtain the Chinese version, we basically follow the instructions of RoBERTa~\cite{DBLP:journals/corr/abs-1907-11692/roberta}.
Considering the Chinese grammar, we adopt WWM in MLM and pre-train on the WuDao Corpora (180 GB version).

The RoBERTa has the following list:
\begin{itemize}
    \item Erlangshen-Roberta-110M-NLI
    \item Erlangshen-Roberta-110M-Sentiment
    \item Erlangshen-Roberta-110M-Similarity
    \item Erlangshen-Roberta-330M-NLI
    \item Erlangshen-Roberta-330M-Sentiment
    \item Erlangshen-Roberta-330M-Similarity
\end{itemize}

\subsubsection{DeBERTa}

We mostly follow the instructions of DeBERTa-v2~\cite{DBLP:conf/iclr/HeLGC21/deberta} to obtain several Chinese versions. 
Considering the Chinese grammar, we adopt WWM in MLM and pre-train on WuDao Corpora (180 GB version) like Erlangshen-MegatronBert.

\begin{itemize}
    \item Erlangshen-DeBERTa-v2-186M-Chinese-SentencePiece
    \item Erlangshen-DeBERTa-v2-320M-Chinese
    \item Erlangshen-DeBERTa-v2-710M-Chinese
    \item Erlangshen-DeBERTa-v2-97M-CWS-Chinese
    \item Erlangshen-DeBERTa-v2-97M-Chinese
\end{itemize}

\subsubsection{Longformer}

By following Longformer~\cite{DBLP:journals/corr/abs-2004-05150/longformer}, we adopt WWM in MLM and pre-train on WuDao Corpora (180 GB version). 
Specifically, we adopt the Rotary Position Embedding (RoPE)~\cite{DBLP:journals/corr/abs-2104-09864/roformer} to avoid the uneven sequence length of the pre-trained corpus.

\begin{itemize}
    \item Erlangshen-Longformer-110M
    \item Erlangshen-Longformer-330M
\end{itemize}

\subsubsection{UBERT}

UBERT~\cite{DBLP:journals/corr/abs-2206-12094/ubert} was the winner solution in the 2022 AIWIN ARTIFICIAL INTELLIGENCE WORLD INNOVATIONS: Chinese Insurance Small Sample Multi-Task\footnotemark[2]. 
Our team, Fengshenbang, developed a unified framework based on BERT for multiple tasks and objectives.
Our UBERT owns first place, as described in leaderboards (A and B).
In addition to the unavailable datasets in the challenge, we carefully collect over $70$ datasets from a variety of tasks for open-source UBERT.
Besides out-of-the-box functionality, our UBERT can be employed in various scenarios such as NLI, entity recognition, and reading comprehension.

\footnotetext[2]{\url{http://ailab.aiwin.org.cn/competitions/68}}

\begin{itemize}
    \item Erlangshen-Ubert-110M-Chinese
    \item Erlangshen-Ubert-330M-Chinese
\end{itemize}

\subsubsection{UnifiedMC (temporary)}

We consider a new paradigm to employ the encoder-based pre-trained language models in zero-shot and few-shot scenarios.
Our code and details about pre-training tasks will be made publicly available upon acceptance of the paper.
In additional, our UnifiedMC models topped the FewCLUE and ZeroCLUE on August 30, 2022.


\subsection{Wenzhong (NLG)}
\label{sec:wenzhong}

The Wenzhong series focus on solving NLG tasks.

\subsubsection{GPT2}

We implement our GPT with $30$ layers to obtain the powerful performance of Chinese GPT2~\cite{radford2019language/gpt2}, which is larger than the original GPT2.
Wenzhong-GPT2-3.5B is pre-trained on CLURCorpus2020~\cite{DBLP:conf/coling/XuHZLCLXSYYTDLS20/clue}. 
The structure of Wenzhong2.0-GPT2-3.5B-chinese is the same as Wenzhong-GPT2-3.5B, and is pre-trained on Wudao Corpus (300G version). 
Moreover, we implement a base size Wenzhong-GPT2-110M with $12$ layers, which is pre-trained on Wudao Corpus (300G version). 

\begin{itemize}
    \item Wenzhong-GPT2-110M
    \item Wenzhong-GPT2-3.5B
    \item Wenzhong2.0-GPT2-3.5B-chinese
\end{itemize}


\subsection{Randeng (NLT)}
\label{sec:randeng}

To handle NLT tasks, we introduce the Randeng series, which aims to solve source-to-target transformation tasks such as text summarization tasks.

\subsubsection{BART}

Based on BART~\cite{DBLP:conf/acl/BART}, we apply BERT tokenizer and WuDao Corpora (180 GB version) to train a Chinese version.
Since the BERT tokenizer usually performs better than others for Chinese tasks, we employ it in one of Randeng-BART models.

\begin{itemize}
    \item Randeng-BART-139M
    \item Randeng-BART-139M-SUMMARY
    \item Randeng-BART-759M-Chinese-BertTokenizer
\end{itemize}

\subsubsection{MegatronT5}

To get a large-scale T5~\cite{DBLP:journals/jmlr/T5}, we make use of Megatron-LM~\cite{DBLP:journals/corr/abs-1909-08053/Megatronlm} method and  WuDao Corpora (180 GB version) for pre-training.

\begin{itemize}
    \item Randeng-MegatronT5-770M
\end{itemize}

\subsubsection{PEGASUS}

To solve Chinese text summarization tasks, we follow the PEGASUS~\cite{DBLP:conf/icml/PEGASUS} guidelines.
We employ the WuDao Corpora (180 GB version) as a pre-training dataset.
In addition, considering that the Chinese sentence piece tokenization is unstable, we utilize jieba\footnotemark[3] and BERT tokenizer in our Randeng-PEGASUS.

\footnotetext[3]{\url{https://github.com/fxsjy/jieba}}

\begin{itemize}
    \item Randeng-Pegasus-238M-Chinese
    \item Randeng-Pegasus-238M-Summary-Chinese
    \item Randeng-Pegasus-523M-Chinese
    \item Randeng-Pegasus-523M-Summary-Chinese
\end{itemize}

\subsubsection{mT5}

We implement mT5~\cite{DBLP:conf/naacl/XueCRKASBR21/mt5} for Chinese. 
In order to accelerate training, we only retrain the vocabulary and embedding corresponding to Chinese and English, and Corpus-Adaptive Pre-Training (CAPT) on the WuDao Corpora (180 GB version).

\begin{itemize}
    \item Randeng-T5-77M
    \item Randeng-T5-784M
\end{itemize}

\subsubsection{Transformer-Denoise}

We explore a Chinese Transformer model to solve denoise tasks.
We first pre-trained Transformer-XL on the Wudo corpus (180G version), and then fine-tuned it on a denoised dataset (developed by us).
The denoise task is to reconstruct a fluent and clean text from a noisy input which includes random insertion/swap/deletion/replacement/sentence reordering.

\begin{itemize}
    \item Randeng-Transformer-1.1B-Denoise
\end{itemize}


\subsection{Taiyi (MM)}
\label{sec:taiyi}

To handle multiple modalities, the Taiyi series is introduced and applied in cross-modal scenarios.
In detail, we are working on protein structure prediction, speech-text representations and so on.

\subsubsection{CLIP}

We follow the experimental setup of CLIP~\cite{DBLP:conf/icml/RadfordKHRGASAM21/clip} to obtain powerful visual-language intelligence.
To obtain the CLIP for Chinese, we employ chinese-roberta-wwm~\cite{DBLP:journals/taslp/CuiCLQY21/wwm} for the language encoder, and apply the ViT in CLIP for the vision encoder. 
We freeze the vision encoder and tune the language encoder to speed up and stabilize the pre-training process.
Moreover, we apply Noah-Wukong dataset~\cite{gu2022/wukong} and Zero-Corpus~\cite{xie2022/zero} as the pre-training datasets.
To the best of our knowledge, our Taiyi-CLIP is currently the only open-sourced Chinese CLIP in the huggingface community.

Moreover, we apply the Taiyi-CLIP models in various text-to-image generation scenarios, which shows the powerful capability of Chinese language understanding, such as generating images with Chinese ancient poems.

\begin{itemize}
    \item Taiyi-CLIP-RoBERTa-326M-ViT-H-Chinese
    \item Taiyi-CLIP-Roberta-102M-Chinese
    \item Taiyi-CLIP-Roberta-large-326M-Chinese
\end{itemize}

\subsubsection{MAP (temporary)}

We propose an exploratory multimodal model to obtain powerful embeddings.
We design a special module to account for multimodal uncertainty and then pre-train the model with special pre-training strategies.
The fine-tuned models are applied to challenging downstream tasks and achieve state-of-the-art performance.
Our framework also helps uni-modal downstream tasks.
Our code and details about pre-training tasks will be made publicly available upon acceptance of the paper.

\begin{itemize}
    \item Taiyi-Roberta-124M-D
    \item Taiyi-Roberta-124M-D-v2
    \item Taiyi-vit-87M-D
\end{itemize}


\subsection{Yuyuan (Domain)}
\label{sec:yuyuan}

Despite being open-sourced, several models still have some difficult issues to solve. 
It is necessary to improve the quality and quantity of existing open source domain-specific datasets. 
But domain-specific model design differs from general PLMs, for example processing proper names. 
For example, for existing models in finance such as FinBERT~\cite{DBLP:journals/corr/abs-2006-08097/finbert}, the vocabularies of it and general BERT~\cite{DBLP:conf/naacl/DevlinCLT19/bert} share $41\%$ common tokens. 
Because of the lack of domain-specific terms in the vocabulary, many domain-specific words are incorrectly represented, damaging the learning of the model. 
Therefore, it is crucial to provide high-quality domain-specific PLMs.

\subsubsection{BioBART}

For the biomedical domain, we adopt BART as presented in our BioBART paper~\cite{DBLP:conf/bionlp/YuanYGZXY22/biobart}.
With Doamin-Adaptive Pre-Training (DAPT), we employ BART-large on PubMed abstracts\footnotemark[4], which contains around 41 GB of biomedical research paper abstracts. 
Next, we collect various biomedical language generation tasks including dialogue, summarization, entity linking, and named entity recognition. 
We demonstrate that BioBART presents large improvements on different benchmarks and achieves competitive or superior results over existing state-of-the-art methods.

\footnotetext[4]{\url{https://pubmed.ncbi.nlm.nih.gov/}; Accessed in 2021.}

\begin{itemize}
    \item Yuyuan-Bart-139M
    \item Yuyuan-Bart-400M
\end{itemize}

\subsubsection{GPT2}

We adopt the same architecture as Wenzhong-GPT2-3.5B to be pre-trained on $50$ GB medical (PubMed) corpus.
Our Yuyuan-GPT2-3.5B is the largest open-source GPT2 model in the medical domain.
We further allow the model to judge facts by computing perplexity (PPL).
To accomplish question-and-answer functionality, we transform the phrase pattern from interrogative to declarative.

\begin{itemize}
    \item Yuyuan-GPT2-3.5B
    \item YuyuanQA-GPT2-3.5B
\end{itemize}


\subsection{TBD (Exploration)}
\label{sec:tbd}

We will have some experimental explorations with several organizations. 
For example, the following models are the outcome of our joint efforts with Zhuiyi Technology.

\begin{itemize}
    \item Zhouwenwang-Unified-1.3B
    \item Zhouwenwang-Unified-110M
\end{itemize}

\section{Fengshen Framework}
\label{sec:framework}

To address the issues in Section~\ref{sec:intro}, we integrate the advantages of Huggingface~\cite{DBLP:journals/corr/abs-1910-03771/huggingface}, Megatron-LM~\cite{DBLP:journals/corr/abs-1909-08053/Megatronlm}, PyTorch-Lightning, and DeepSpeed~\cite{DBLP:conf/sc/RajbhandariRRH20/deepspeed} into Fengshen framework.
If the users are familiar with the above frameworks, then they can use our deep learning framework without any effort.
Our framework can be utilized to pre-train large-scale models even over $10$ billion parameters by using terabytes level data and be fine-tuned a range of downstream tasks.
With this configuration, users can simply undertake distributed training and memory saving approaches, allowing them to focus more on model implementation and innovation.
The Fengshen framework also provides codes and examples for open-source models in Huggingface and their applications.

Our framework has the following advantages:
\begin{itemize}
    \item Superior performance compared to the original Torch, such as $300\%$ acceleration in training.
    \item Support large-scale models: over $10$ billion-level models in training and fine-tuning.
    \item Large datasets (Terabyte-level) are supported.
    \item The training process is easy to use by providing rich pre-training and downstream examples.
    \item Adapt to varied device environments (various devices such as CPU, GPU, and TPU).
    \item Support distributed training methods such as DDP and Zero Optimizer without any code modifications.
\end{itemize}

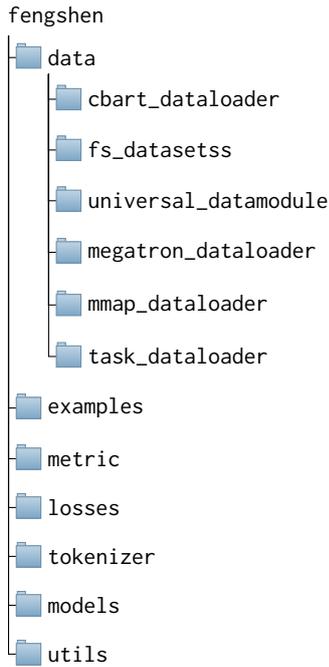
\begin{figure}
\small
\begin{forest}
  for tree={
    font=\ttfamily,
    grow'=0,
    child anchor=west,
    parent anchor=south,
    anchor=west,
    calign=first,
    inner xsep=7pt,
    edge path={
      \noexpand\path [draw, \forestoption{edge}]
      (!u.south west) +(7.5pt,0) |- (.child anchor) pic {folder} \forestoption{edge label};
    },
    before typesetting nodes={
      if n=1
        {insert before={[,phantom]}}
        {}
    },
    fit=band,
    before computing xy={l=15pt},
  }  
[fengshen
  [data
    [cbart\_dataloader]
    [fs\_datasetss]
    [universal\_datamodule]
    [megatron\_dataloader]
    [mmap\_dataloader]
    [task\_dataloader]
  ]
  [examples
  ]
  [metric
  ]
  [losses
  ]
  [tokenizer
  ]
  [models
  ]
  [utils
  ]
]
\end{forest}
\caption{An overview of Fengshen Framework.}
\label{fig:file_tree}
\end{figure}


\begin{figure*}
\centering
\includegraphics[width=\textwidth]{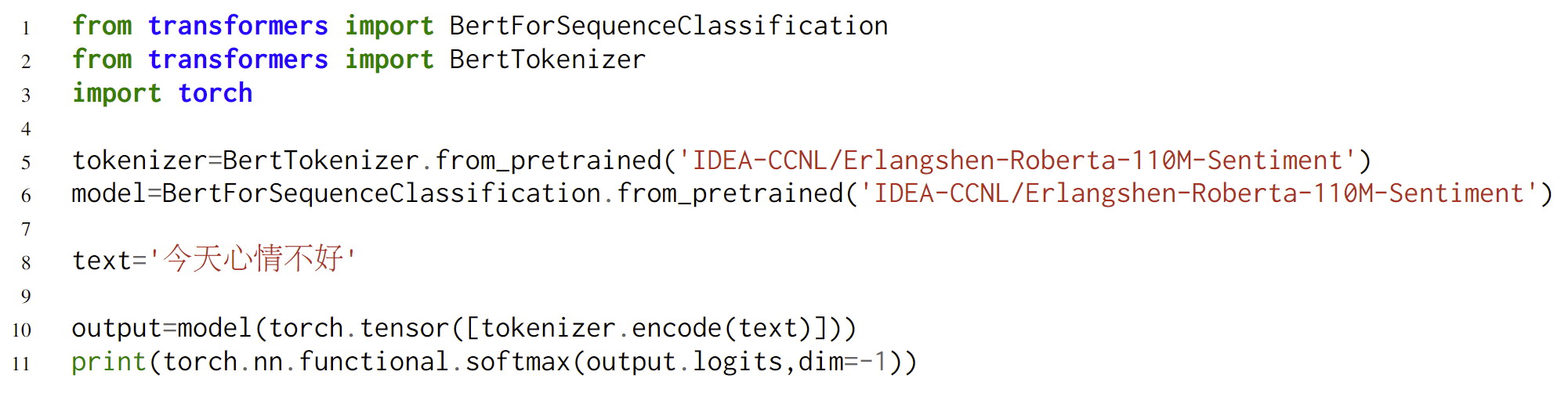}
\caption{An example for generating task with Erlangshen-Roberta-110M-Sentiment.}
\label{fig:code_example}
\end{figure*}

\subsection{Architecture}

Core components in the Fengshen Framework are presented in Figure~\ref{fig:file_tree}:
The core functionality of the model is to adapt to each module.
It is straightforward to follow 3 steps to apply our Fengshen Framework:

\begin{enumerate}
    \item Encapsulated Data Processing Flow
    \item Encapsulate the model structure
    \item Configure several plugins
\end{enumerate}

\subsection{Document}

This paper provides the usage details\footnotemark[5]~ of distributed model training, fine-tuning, and various large scale model applications.
Then, we present the paper and various real-world code and tutorials from our participation competition. 

\footnotetext[5]{\url{https://fengshenbang-doc.readthedocs.io/zh/latest/index.html}}

\subsection{Case Study}

As shown in Fig.~\ref{fig:code_example}, only a few lines are required to construct the models in the Fengshen framework.
For users who are already using frameworks like Huggingface, there is almost no cost to get started.

\section{Fengshenbang Benchmark}
\label{sec:benchmark}

Benchmarks are necessary to evaluate the capability of the models.
It should be noted that Chinese is linguistically distinct from English and other Indo-European languages.
Thus, we introduce Fengshenbang Benchmarks, which are composed of Chinese leaderboard system with different types of natural language tasks.
Besides considering fair environments in several tasks, we plan to release a general pre-training dataset.

\subsection{Task Criteria}

To collect high-quality and robust benchmarks, we consider different aspects of testing the models.
As a result, we identify the following requirements while building the Fengshenbang benchmark:

\noindent\textbf{Widely evaluated.} While some existing datasets are not designed in Chinese, they have been used extensively in NLP for years, e.g. SuperGLUE~\cite{DBLP:conf/nips/WangPNSMHLB19/superglue}. 
We will gather some professional English and Chinese linguists to meticulously translate these popular datasets.

\noindent\textbf{Future-oriented.} In fact, a few NLP models already surpass human performance on several benchmarks.
This declares that AI has reached or even can surpass human cognitive intelligence.
One reason we believe is their limited scope of evaluation.
A more urgent and necessary work is to construct challenging datasets instead of fitting existing datasets to $100\%$ accuracy.
Future benchmarks need to consider broader ethical, technical, and societal challenges.
Our datasets will be published soon to better support the research community.

\noindent\textbf{Applicable.} Benchmarks are required to represent real-world scenarios.
This allows us to collaborate with industry-active companies to publish datasets and collect real-world data.

\subsection{Leaderboard}

\subsubsection{Chinese-SuperGLUE}

As a powerful benchmark, the SuperGLUE is widely adopted, but it is only available in English.
These high-quality datasets are hoped for by the Chinese community as well.
The Chinese-language-oriented model can be challenged due to the lack of validation on the English dataset.
This is very unfair and may result in the research being limited to a monolingual setting.
One reason is that it is difficult to evaluate the quality of different datasets, leading reviewers to be more surprised about some unseen datasets.
To address the above issues, we develop Chinese-SuperGLUE to evaluate Chinese models.
\textit{The related paper and the leaderboard are coming soon.}

\subsubsection{QAKM}

To evaluate the knowledge level in NLP models, we propose QAKM (Question Answering with Knowledge Models).
The knowledge models are required to learn domain-specific knowledge and answer unseen questions given a dataset.
The task has been included in the NLPCC\footnotemark[6], which is accessible in this website\footnotemark[7].


\footnotetext[6]{\url{http://tcci.ccf.org.cn/conference/2022/cfpt.php}}
\footnotetext[7]{\url{https://idea.edu.cn/ccnl-act/nlpcc-track1.html}}

\section{Summary and Future Work}
\label{sec:future}

This report introduces our open source project Fengshenbang, aiming to build the foundation of Chinese Cognitive Intelligence.
Our three sub-projects (Fengshenbang Model, Fengshen Framework, Fengshenbang Benchmark) support different aspects of intelligence systems' progress.
In addition, we want to emphasize that our project Fengshenbang is ongoing, i.e., we keep all sub-projects updated forward.
When building a community, we expect individual and organizational contributors to join and refine the project together.
The world needs a few good ideas.

\section*{Ethical Considerations}
\label{sec:ethical}


However, since our Fengshenbang Project provides the entire ecosystem to use, produce, and evaluate large-scale PLMs, we note that many companies and research institutions have deployed these models.
Our models and benchmarks transfer gradually to the real world, thus having unpredictable impact on humans.
There are many aspects to be concerned with considering the ethical impact: implicit bias of large-scale models, potential environmental issues, undesirable prejudgement in labeled data, inappropriate use of open source frameworks, etc.
For a better understanding and deeper discussion of ethical issues, we will provide a detailed study in the next version of the report.
While we encourage any developer to use anything from the Fengshenbang project to open debate on its utilization, such as task selection and deployment, we hope that this would reduce the chance of any misconduct.

\section*{Acknowledgements}
We thank the CCNL GTS team in IDEA for providing feedback and help.

\clearpage

\section{引言}
\label{sec:intro_zh}

人工智能~(Artificial Intelligence, AI)的显著进步产生了许多伟大的模型，特别是基于预训练的基础模型~\cite{DBLP:journals/corr/abs-2108-07258/opportunities}成为了一种新兴的范式。
传统的AI模型必须要在专门的巨大的数据集上为一个或几个有限的场景进行训练，相比之下，基础模型可以适应广泛的下游任务。
因此，基础模型让低资源的场景有AI落地的可能。
并且，我们观察到这些模型的参数量正在以每年10倍的速度增长。
比如，于2018年，BERT~\cite{DBLP:conf/naacl/DevlinCLT19/bert}的参数量仅有1亿量级。
但是到了2020年，GPT-3~\cite{DBLP:conf/nips/BrownMRSKDNSSAA20/gpt3}的参数量就已达到百亿的量级。
由于这一鼓舞人心的趋势，人工智能中的许多前沿挑战，尤其是强大的泛化能力，正在变得可以被实现。

基础模型，尤其是语言模型，如今正在被英文社区主导着。
然而，中文作为这个世界上最大的口语语种（母语者中），却缺乏系统性的研究资源支撑，这使得中文领域的研究进展相较于英文来说有些滞后。
为了解决中文领域研究进展滞后和研究资源严重不足的问题，我们提出了一个名为``封神榜''的中文驱动的基础生态系统，其中包括了预训练模型，特定任务的微调应用，基准和数据集等。

\begin{figure}[tbp]
  \centering
  \includegraphics[width=0.5\textwidth]{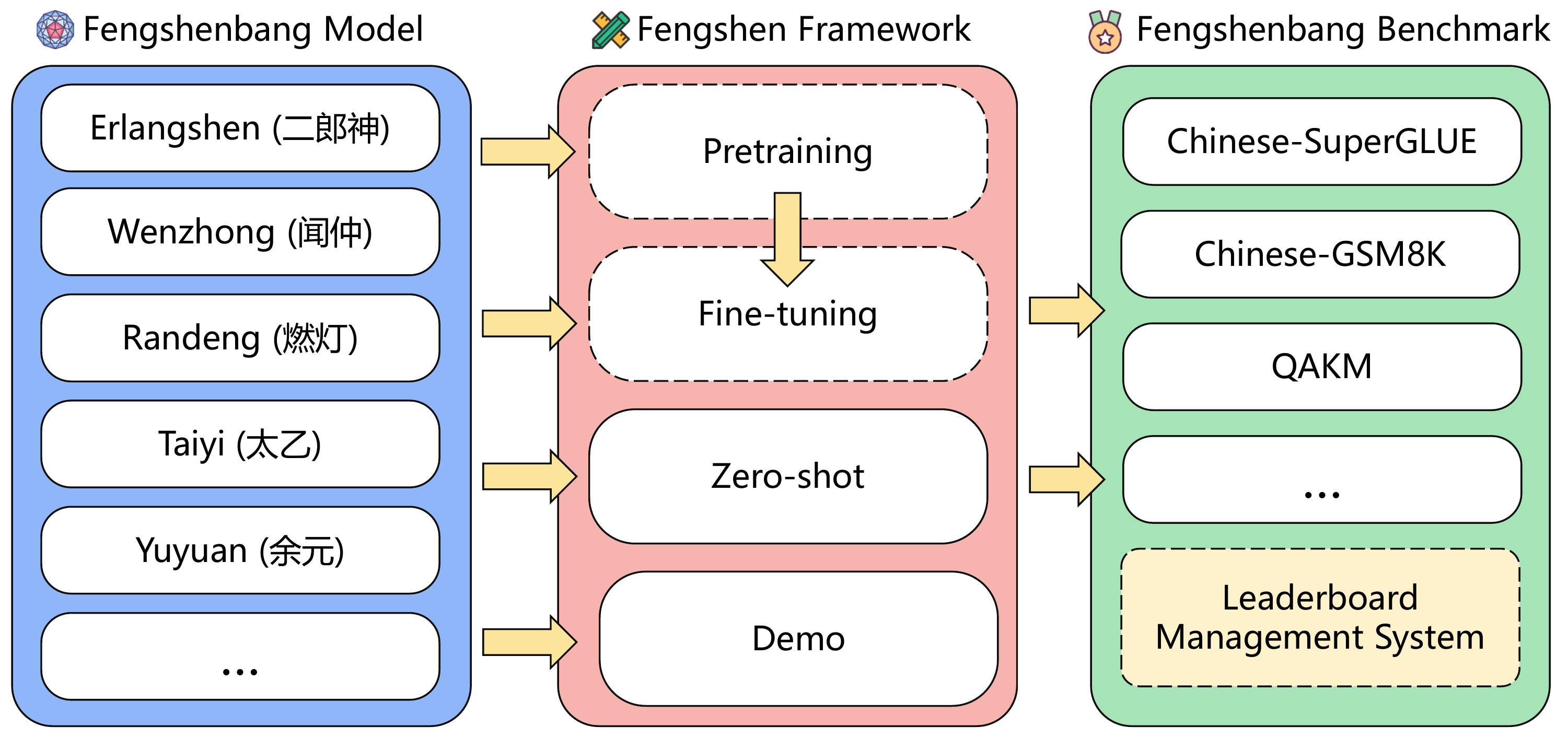}
  \caption{封神榜项目概览。}
  \label{fig:introduction_zh}
\end{figure}

我们的目标是构建一个全面的，标准化的，以用户为中心的生态系统。
尽管这一目标可以通过多种方式去实现，但是我们经过对中文社区的重新审视与思考，提出了我们认为最为有效的方案。
具体地，沈向洋在IDEA大会上正式宣布启动``封神榜''项目，旨在成为中文认知智能的基础设施。
并且，我们希望通过全面开放和协作的方式共建开源社区，并且推动中文自然语言技术发展。
如图\ref{fig:introduction_zh}所示，封神榜包括了三个核心模块：
\begin{itemize}
    \item 封神榜模型 (第\ref{sec:model_zh}节)
    \item 封神框架 (第\ref{sec:framework_zh}节)
    \item 封神榜单 (第\ref{sec:benchmark_zh}节)
\end{itemize}

在具体讨论封神榜模型之前，我们先引入一些必要的概念。
NLP社区中有着广泛的研究任务，这些任务可以被分为两类：通用任务和特殊任务。
前者包括了自然语言理解(NLU)，自然语言生成(NLG)和自然语言转换(NLT)任务。
后者涵盖了多模态，特定领域等任务。
重要的是，我们认为一个综合的基础模型平台是有必要的，因此，我们在封神榜中考虑了上述所有任务。
此外，我们还提供了在下游任务上微调好的相关模型，这使得计算资源有限的用户也可以轻松使用我们的基础模型。

接着，我们引入以用户为中心的封神框架。用户可以根据我们提供的资源进一步完善或者修改模型。
具体地，我们提供了简单灵活的使用方案，用户可以低成本地使用：包含标准化的数据处理器、详细的教程示例、类docker的环境和行业标准的API等。

最后，我们提出的生态系统中还包括了一个基准模块，它允许用户进行公平的比较并且可以让整个社区追踪最新进展。
特别地，我们在未来会发布榜单管理系统，希望可以推动更多定制的排行榜系统的发展。

以上三个模块就是我们的整个生态系统了。
尽管这看起来可能有些复杂，但是只需三步，用户就可以根据我们的资源轻松构建所需的应用了。
\begin{enumerate}
    \item[步骤 1:] 从我们的封神榜模型库中选择一个预训练好的中文NLP模型.
    \item[步骤 2:] 通过阅读我们的教程示例，使用封神框架调整模型。
    \item[步骤 3:] 评估下游任务，如封神榜单或者自定义任务。
\end{enumerate}

\section{封神榜模型}
\label{sec:model_zh}

在这一节中，我们将会介绍封神榜模型相关的细节。
如表\ref{table:user_tax_zh}所示，我们展示了一个以用户为中心的分类学(User-Centered Taxonomy, UCT)。
UCT会专门对用户的需求进行分类，提供相对应的模型和相关模型的设计指导。
这些模型更多的细节将会在第\ref{sec:model_selelction_zh}节中说明。
用户可以根据我们的模型选择标准快速的定位到对应自身需求的模型。
事实上，我们还提供了第\ref{sec:naming_convention_zh}节中所叙述的命名法，来尽可能地降低寻找所需模型的难度。
具体的系列说明以及其对应的模型可以在第\ref{sec:erlangshen_zh}节到第\ref{sec:tbd_zh}节中找到。
此外，我们所有的49个模型都可以在附录\ref{append:all_model}中查到。


\subsection{模型设计}

\begin{table}[]
\small
\centering
\begin{tabular}{l|c|c|c}
\toprule
需求                   & 任务        &系列     & 示例模型 \\ \midrule
\multirow{3}{*}{通用} & 自然语言理解        & 二郎神 & BERT, DeBERTa          \\
                         & 自然语言生成        & 闻仲   & GPT2          \\
                         & 自然语言转换        & 燃灯    & T5, BART            \\ \midrule
\multirow{3}{*}{特定} & 多模态          & 太乙     & CLIP          \\
                         & 领域     & 余元     & BioBART    \\
                         & 探索 & 待定  & 待定      \\ \bottomrule
\end{tabular}
\caption{以用户为中心的分类学和模型示例。}
\label{table:user_tax_zh}
\end{table}


\subsubsection{以用户为中心的分类学(UCT)}
\label{sec:model_taxonomy_zh}

NLP的先进研究中已经出现了大量从不同角度(研究和应用等)的强大模型。
为了理解模型的不同以及跟踪社区的进展，我们希望可以建立一个应用于NLP领域的标准的分类法。
然而，因为模型通常由于其复杂性而难以分类。
比如，TinyBERT~\cite{DBLP:conf/emnlp/JiaoYSJCL0L20/tinybert}在模型构架中可以被归为``Encoder-only''，也可以在模型参数缩减方法中归为``Distillation''。
为了减少误解，我们引入了以用户为中心的分类学(UCT)，它来自于众多NLPer的意见和咨询。现有的需求通常可以分为通用需求和特殊需求。
在通用需求中，有常见的NLP任务，其被分为自然语言理解(NLU)、自然语言生成(NLG)和自然语言转换(NLT)任务。
由于快速的发展，NLP社区给整个AI社区带来了特殊的需求，这些需求涵盖了多模态，特定领域和探索等。
这个分类并不会止步不前，而是会随着用户需求的变化而与时俱进。

\noindent{\ul 自然语言理解(NLU)}

NLU任务要求模型利用文本的句法和语义进行分析来理解句子的含义。
句法和句子的语法结构相关，语义则代表了其预期含义。
单词和短语之间的关系也很重要，因为它们会导致概念理解上的差异。
此外，这项任务中的一些问题即使是人类也难以解决。
为了评估模型的NLU能力，如下的下游任务被设计出来以检验模型的可靠性：

\begin{itemize}
    \item 语义匹配
    \item 情感分析
    \item 自然语言推理
    \item 实体识别
    \item 关系抽取
    \item 事件抽取
    \item 中文分词
    \item \dots
\end{itemize}

\noindent{\ul 自然语言生成 (NLG)}

与NLU中的机器阅读理解能力不同，NLG关注于开发能够创作可被人类理解的写作内容的的计算系统。
具有NLG能力的系统应该能够通过自身或所学的思想来生成自然语言，而不是转换现有的数据形式。
而且，生成的文本需要连贯且易于理解。
一些NLG任务如下：
\begin{itemize}
    \item 创意写作
    \item 因果推理
    \item 可控生成
    \item 多步推理
    \item \dots
\end{itemize}

\noindent{\ul 自然语言转换 (NLT)}

NLT任务是我们定义的一种全新的任务类型，是一种源→目标的转换任务。不同于NLG基于自己思想的生成，NLT要求语言模型在理解源文本对象的基础上，生成或者转换出目标文本对象。
以机器翻译为例，给定一种语言的文本，AI系统需要生成另一种语言的相应文本。
总之，我们将部分NLT任务在下面列出。

\begin{itemize}
    \item 机器翻译
    \item 文本摘要
    \item 文本简化
    \item 语法纠错
    \item 问答系统
    \item 对话系统
    \item \dots
\end{itemize}

\noindent{\ul 多模态 (MM)}

如今，复杂场景的需求在不断增长，而单模态的模型无法处理这些场景中的多模态信息。
基于Transformer的预训练语言模型(Pre-trained Language Models, PLMs)因其灵活的架构而广泛用于计算机视觉和音频处理等领域。
此外，认知智能需要智能系统从多种模式中学习，包括文本、图像和音频。
为此，我们引入了多个多模态场景，如文本生成图像和多模态语义理解等。
一些多模态的任务如下：

\begin{itemize}
    \item 文本生成图像
    \item 图像标注
    \item 跨模态检索
    \item 视觉问答
    \item 自动语音识别
    \item 文字转语音
    \item 语音会话
    \item 蛋白质结构预测
    \item \dots
\end{itemize}

\noindent{\ul 领域模型 (Domain)}

值得注意的是，PLMs在各个特定领域中都取得了惊人的成功。
持续的预训练是特定领域模型的关键优势。
因为模型不是从头开始预训练的，所以计算资源消耗更少。
下面列出了一些领域和几个相关模型：

\begin{itemize}
    \item 金融: FinBERT~\cite{DBLP:journals/corr/abs-2006-08097/finbert}
    \item 生物医学: BioBERT~\cite{DBLP:journals/bioinformatics/LeeYKKKSK20/biobert}, ClinicalBERT~\cite{DBLP:journals/corr/abs-1904-03323/clinical_bert}, PubMedBERT~\cite{DBLP:journals/health/GuTCLULNGP22/pubmed_bert}
    \item 法律: LEGAL-BERT~\cite{DBLP:journals/corr/abs-2010-02559/legalbert}, ALeaseBERT~\cite{DBLP:journals/corr/abs-2010-10386/aleasebert}
    \item 编程: CoTexT~\cite{DBLP:journals/corr/abs-2105-08645/cotext}, CodeBERT~\cite{DBLP:conf/emnlp/FengGTDFGS0LJZ20/codebert}, GraphCodeBERT~\cite{DBLP:conf/iclr/GuoRLFT0ZDSFTDC21/graphcodebert}, CodeGPT-adapted~\cite{DBLP:conf/nips/LuGRHSBCDJTLZSZ21/codegpt-adap}, Codex~\cite{DBLP:journals/corr/abs-2107-03374/codex}
    \item 学术: OAG-BERT~\cite{DBLP:journals/corr/abs-2103-02410/oagbert}, MathBERT~\cite{DBLP:journals/corr/abs-2105-00377/mathbert}, SciBERT~\cite{DBLP:conf/emnlp/BeltagyLC19/scibert}
    \item \dots
\end{itemize}

\noindent{\ul 探索}

我们也希望和其他组织，如技术公司和大学，一起开发一些关于NLP的实验性模型。

\subsubsection{模型选择}
\label{sec:model_selelction_zh}

由于每年都有许多论文和模型被提出，为了控制整体质量和合理使用计算资源，我们只能选择其中的一部分进行预训练，然后进行开源。
所以，我们根据以下规则来选择模型：

\noindent{\ul 强大的。} 
一些模型在下游任务上呈现出惊人的性能。
但是，这些模型要么是以英文发表，要么甚至没有发布出来。
此外，它们往往与中文社区不匹配，但需要扩展或修改。

\noindent{\ul 多样性。} 
有很多模型已经被广泛得应用于多种多样的NLP任务中了，比如BERT~\cite{DBLP:conf/naacl/DevlinCLT19/bert}, GPT-3~\cite{DBLP:conf/nips/BrownMRSKDNSSAA20/gpt3}和Transformer~\cite{DBLP:conf/nips/VaswaniSPUJGKP17/transformer}. 
并且，他们通常易于拓展和调整。
我们收纳这些模型的时候，会考虑到下游任务、架构、模型大小和预训练方法等不同情况。

\noindent{\ul 实用的。} 
开放源码的模型应该容易被理解并且是可以实现的。
此外，用户可以开箱即用地使用模型于所需的下游任务。


\subsubsection{命名规则}
\label{sec:naming_convention_zh}

为了更好地理解封神榜模型，我们引入了一个新的命名空间，模板如下： 
\begin{equation} 
\small 
Name \in \{ Series-Model-Parameter-Extra \}^N 
\end{equation} 
其中，$Series$ 是取自中国小说《封神榜》中神话人物的名字~\cite{hsun2000brief/fengshenbang}。
每个系列对应着一类NLP任务。
$Model$ 是模型的结构，例如 BERT 和 GPT-2。
$Parameter$ 是参数的数量，$N$ 是我们开源模型的总数。
$Extra$ 是额外的信息，例如，在特定下游任务的数据集上微调过。
代表了二郎神，给出一个例子，``Erlangshen-Roberta-110M-NLI''代表了这是二郎神系列，用于解决自然语言理解(NLU)任务的。
模型结构基于具有110M参数量的RoBERTa模型。
在对中文数据集进行预训练后，我们继续在自然语言推理(NLI)任务上对其进行微调。


\subsection{二郎神 (NLU)}
\label{sec:erlangshen_zh}

二郎神系列旨在解决NLU问题，目前包括Megatron-BERT, ZEN, RoBERTa, DeBERTa, Longformer, UBERT和UnifiedMC等模型。

\subsubsection{MegatronBERT}

为了训练十亿级规模的BERT，我们参考Megatron-LM~\cite{DBLP:journals/corr/abs-1909-08053/Megatronlm}的方法，在悟道语料库(180 GB 版本)~\cite{ DBLP:journals/aiopen/YuanZDDLCZYT21/wudao_corpora}上预训练BERT~\cite{DBLP:conf/naacl/DevlinCLT19/bert}。
鉴于中文语法结构和大规模模型训练的难度，我们应用以下四种预训练策略来改进BERT~\cite{DBLP:conf/naacl/DevlinCLT19/bert}.\\ 
(1) 整词掩码(Whole Word Masking, WWM)。
考虑到中文的语言特性，我们采用WWM~\cite{DBLP:journals/taslp/CuiCLQY21/wwm}，即在WordPiece分词器中处理整个中文单词而不是单个汉字。\\ 
(2) 基于知识的动态掩码 (Knowledge-based Dynamic Masking, KDM)。
替换原本的掩码语言建模 (MLM) 中的随机掩码，我们尝试去遮掩那些具有更丰富语义的信息以产生有效和高效的模型。\\ 
(3) 句序预测 (Sentence Order Prediction, SOP)。
参考ALBERT~\cite{DBLP:conf/iclr/LanCGGSS20/albert}，语言模型可以学习句间信息以获得强大的表征。
因此，Erlangshen-MegatronBert模型使用SOP而不是NSP。\\ 
(4) 前层归一化 (Pre-LN)。
在语言模型预训练阶段使用层后归一化时，我们发现随着模型大小的增加，损失会异常上升的问题。
因此，我们应用Pre-LN~\cite{DBLP:conf/icml/XiongYHZZXZLWL20/pre-ln} 来克服这个问题。

据我们所知，在我们开源rlangshen-MegatronBert-1.3B的时候，它是当时中国开源社区中参数量最大的BERT模型。
Erlangshen-MegatronBert在Hugging Face上有着每月4.5K\footnotemark[8] 的下载量，帮助了许多的开发人员。

\footnotetext[8]{数据获取时间为2022年8月18日。}

因为其出色的表现，Erlangshen-MegatronBert获得了三个重要的成就：\\ 
(1) 2021 年 11 月 10 日，它在CLUE基准~\cite{DBLP:conf/coling/XuHZLCLXSYYTDLS20/clue}的FewCLUE上取得第一。
其中，它在CHIDF(成语填空)和TNEWS(新闻分类)子任务中的表现优于人类表现。
此外，它在CHIDF(成语填空), CSLDCP(学科文献分类), OCNLI(自然语言推理)任务中均名列前茅。\\
(2) 2022年1月24日，它在CLUE基准测试中的ZeroCLUE中取得第一。
对于这些任务中的每一个，我们在 CSLDCP(主题文献分类), TNEWS(新闻分类), IFLYTEK(应用描述分类), CSL(抽象关键字识别)和 CLUEWSC(指代消歧)任务中取得第一。\\
(3) 在2022年7月10日，我们在CLUE基准的语义匹配任务中取得第一~\cite{DBLP:journals/corr/abs-2208-02959/no1_simclue}。

\begin{itemize}
    \item Erlangshen-MegatronBert-1.3B
    \item Erlangshen-MegatronBert-1.3B-NLI
    \item Erlangshen-MegatronBert-1.3B-Sentiment
    \item Erlangshen-MegatronBert-1.3B-Similarity
    \item Erlangshen-MegatronBert-3.9B-Chinese
\end{itemize}

下一步，我们将公开一篇论文去详细介绍Erlangshen-MegatronBert的相关信息。

\subsubsection{ZEN}

我们与ZEN团队合作，使用我们的封神框架，开源发布了ZEN1~\cite{DBLP:conf/emnlp/DiaoBSZW20/zen1}和ZEN2~\cite{DBLP:journals/corr/abs-2105-01279/zen2}模型。
具体而言，通过引入无监督学习中提取的知识，ZEN1通过N-gram方法学习不同的文本粒度信息。
ZEN1可以通过仅在单个小语料库(低资源场景)上进行训练来获得良好的性能增益。
ZEN2使用大规模数据集和特殊的预训练策略对n-gram增强编码器进行预训练。
下一步，我们将继续与ZEN团队一起探索PLM的优化，并提高下游任务的性能。

\begin{itemize}
    \item Erlangshen-ZEN1-224M-Chinese
    \item Erlangshen-ZEN2-345M-Chinese
    \item Erlangshen-ZEN2-668M-Chinese
\end{itemize}

\subsubsection{RoBERTa}

为了获得其中文版本，我们基本上遵循RoBERTa ~\cite{DBLP:journals/corr/abs-1907-11692/roberta}的做法。
考虑到中文语法的特性，我们在MLM中采用WWM，并在悟道语料库(180 GB版本)上进行预训练。

\begin{itemize}
    \item Erlangshen-Roberta-110M-NLI
    \item Erlangshen-Roberta-110M-Sentiment
    \item Erlangshen-Roberta-110M-Similarity
    \item Erlangshen-Roberta-330M-NLI
    \item Erlangshen-Roberta-330M-Sentiment
    \item Erlangshen-Roberta-330M-Similarity
\end{itemize}

\subsubsection{DeBERTa}

我们主要按照DeBERTa-v2 ~\cite{DBLP:conf/iclr/HeLGC21/deberta}的论文，结合开源的代码来训练我们的中文版本。
考虑到中文语法的特性，我们在MLM中采用WWM，并在悟道语料库(180 GB版本)上进行预训练。

\begin{itemize}
    \item Erlangshen-DeBERTa-v2-186M-Chinese-SentencePiece
    \item Erlangshen-DeBERTa-v2-320M-Chinese
    \item Erlangshen-DeBERTa-v2-710M-Chinese
    \item Erlangshen-DeBERTa-v2-97M-CWS-Chinese
    \item Erlangshen-DeBERTa-v2-97M-Chinese
\end{itemize}

\subsubsection{Longformer}

遵循Longformer~\cite{DBLP:journals/corr/abs-2004-05150/longformer}的设计，我们在MLM中采用WWM，并在悟道语料库(180 GB版本)上进行了中文版Longformer的预训练。
特别的，我们采用旋转位置嵌入(RoPE)~\cite{DBLP:journals/corr/abs-2104-09864/roformer}来避免预训练语料库的不均匀序列长度问题。

\begin{itemize}
    \item Erlangshen-Longformer-110M
    \item Erlangshen-Longformer-330M
\end{itemize}

\subsubsection{UBERT}

UBERT~\cite{DBLP:journals/corr/abs-2206-12094/ubert}是2022年AIWIN世界人工智能创新大赛：中文保险小样本多任务竞赛的冠军解决方案\footnotemark[9]。
我们开发了一个基于BERT的多任务、多目标、统一的抽取任务框架。
我们的UBERT在比赛A榜和B榜上均取得了第一名。
因为比赛中的数据集在比赛结束后不再可用，我们开源的UBERT从多个任务中收集了70多个数据集来进行预训练。
除了支持开箱即用之外，我们的UBERT还可以用于各种场景，如NLI、实体识别和阅读理解。

\footnotetext[9]{\url{http://ailab.aiwin.org.cn/competitions/68}}

\begin{itemize}
    \item Erlangshen-Ubert-110M-Chinese
    \item Erlangshen-Ubert-330M-Chinese
\end{itemize}

\subsubsection{UnifiedMC (暂定)}

我们提出了一种新的范式，即在零样本和少样本场景中使用基于双向编码器的预训练语言模型。
我们的代码和有关预训练任务的详细信息将在论文被接收后公开。
值得注意的是，我们的 UnifiedMC 模型在的FewCLUE和ZeroCLUE中均取得第一(时间：2022年8月30日)。


\subsection{闻仲 (NLG)}
\label{sec:wenzhong_zh}

闻仲系列专注于解决NLG任务。

\subsubsection{GPT2}

我们实现了30层的GPT，来获得当前我们开源中文GPT2~\cite{radford2019language/gpt2}的强大性能，并且它比原始GPT2在参数量上更大。
Wenzhong-GPT2-3.5B在 CLURCorpus2020~\cite{DBLP:conf/coling/XuHZLCLXSYYTDLS20/clue} 上进行预训练。
Wenzhong2.0-GPT2-3.5B-chinese的结构与Wenzhong-GPT2-3.5B相同，不同的是，该版本是在悟道语料库(300G版本)上进行的预训练。
此外，我们实现了一个12层的Wenzhong-GPT2-110M，它也是在悟道语料库(300G版本)上进行预训练的。

\begin{itemize}
    \item Wenzhong-GPT2-110M
    \item Wenzhong-GPT2-3.5B
    \item Wenzhong2.0-GPT2-3.5B-chinese
\end{itemize}


\subsection{燃灯 (NLT)}
\label{sec:randeng_zh}

为了处理NLT任务，我们引入了燃灯系列，即目标是解决源对象到目标对象的转换任务，比如文本摘要任务。

\subsubsection{BART}

基于BART~\cite{DBLP:conf/acl/BART}，我们应用BERT的分词器和悟道语料库(180G版本)训练了一个中文版本。因为BERT分词器通常在中文任务中表现比其他分词器好，所以我们在一个Randeng-BART模型中使用了它。

\begin{itemize}
    \item Randeng-BART-139M
    \item Randeng-BART-139M-SUMMARY
    \item Randeng-BART-759M-Chinese-BertTokenizer
\end{itemize}

\subsubsection{MegatronT5}

为了得到一个大规模的T5~\cite{DBLP:journals/jmlr/T5}，我们使用了Megatron-LM~\cite{DBLP:journals/corr/abs-1909-08053/Megatronlm}的方法和悟道语料库(180G版本)用于预训练。

\begin{itemize}
    \item Randeng-MegatronT5-770M
\end{itemize}

\subsubsection{PEGASUS}

为了解决中文的自动摘要任务，我们遵循PEGASUS~\cite{DBLP:conf/icml/PEGASUS}的设计来训练中文的版本。
我们使用了悟道语料库(180G版本)作为预训练数据集。
此外，考虑到中文sentence piece不稳定，我们在Randeng-PEGASUS中同时使用了结巴分词\footnotemark[10]和BERT分词器。

\footnotetext[10]{\url{https://github.com/fxsjy/jieba}}

\begin{itemize}
    \item Randeng-Pegasus-238M-Chinese
    \item Randeng-Pegasus-238M-Summary-Chinese
    \item Randeng-Pegasus-523M-Chinese
    \item Randeng-Pegasus-523M-Summary-Chinese
\end{itemize}

\subsubsection{mT5}

我们实现了中文版的mT5~\cite{DBLP:conf/naacl/XueCRKASBR21/mt5}。
为了加速训练，我们只在悟道语料库(180G版本)上重新训练了中英文对应的词汇和嵌入，并且使用了语料库自适应预训练(Corpus-Adaptive Pre-Training, CAPT)技术。

\begin{itemize}
    \item Randeng-T5-77M
    \item Randeng-T5-784M
\end{itemize}

\subsubsection{Transformer-Denoise}

我们做了一个基于中文的Transformer模型的塔拉索性实验，希望可以利用其解去噪任务。
我们先使用Transformer-XL的模型结构在悟道语料库(180G版本)上进行预训练，然后在我们自主构建的去噪数据集上进行微调。其中，去噪任务是从包括 随机插入/交换/删除/替换/句子重排 的具有噪声的输入中重建一个流畅和干净的文本。

\begin{itemize}
    \item Randeng-Transformer-1.1B-Denoise
\end{itemize}


\subsection{太乙 (MM)}
\label{sec:taiyi_zh}

为了处理多种模态，我们引入了太乙系列，并将其应用于跨模态场景。
具体来说，我们正在研究蛋白质结构预测, 语音-文本表示等任务。

\subsubsection{CLIP}

我们遵循CLIP~\cite{DBLP:conf/icml/RadfordKHRGASAM21/clip}的实验设置，以获得强大的视觉-语言表征。
在训练中文版的CLIP时，我们使用chinese-roberta-wwm~\cite{DBLP:journals/taslp/CuiCLQY21/wwm}作为语言的编码器，并将CLIP中的ViT应用于视觉的编码器。
为了快速且稳定地进行预训练，我们冻结了视觉编码器并且只微调语言编码器。
此外，我们将Noah-Wukong数据集~\cite{gu2022/wukong}和Zero语料库~\cite{xie2022/zero}用作预训练的数据集。
据我们所知，我们的Taiyi-CLIP是目前Huggingface社区中唯一的开源中文CLIP。

\begin{itemize}
    \item Taiyi-CLIP-RoBERTa-326M-ViT-H-Chinese
    \item Taiyi-CLIP-Roberta-102M-Chinese
    \item Taiyi-CLIP-Roberta-large-326M-Chinese
\end{itemize}

\subsubsection{MAP (暂定)}

我们提出了一个探索性的多模态模型，以获得强大的表征。
我们设计了一个特殊模块来解决多模态的不确定性问题，然后使用特殊的预训练策略对模型进行预训练。
微调后的模型被应用于具有挑战性的多模态下游任务，并且能够实现先进的性能。
我们的框架也有助于单模态的下游任务。
其代码和有关预训练任务的详细信息将在论文被接收后公开。

并且，我们把Taiyi-CLIP模型应用于文本生成图像的场景中，其展示了强大的中文语言理解能力，比如可以利用中文古诗词生成图片。

\begin{itemize}
    \item Taiyi-Roberta-124M-D
    \item Taiyi-Roberta-124M-D-v2
    \item Taiyi-vit-87M-D
\end{itemize}


\subsection{余元 (Domain)}
\label{sec:yuyuan_zh}

尽管我们开源了很多解决通用需求的模型，但是在一些特定的领域中仍然有一些困难的问题尚未被解决。
但领域特定的模型设计不同于一般的预训练语言模型，例如需要处理专有名称等。
作为例子，在金融领域中，现有的模型如 FinBERT~\cite{DBLP:journals/corr/abs-2006-08097/finbert}，它的词汇表和通用的 BERT~\cite{DBLP:conf/naacl/DevlinCLT19/bert}仅仅共享$41\%$的token。
由于词汇表中缺少特定领域的术语，许多特定领域的词被错误地表示，从而损害了模型的学习。
因此，我们认为提高现有领域特定的开源模型的质量和数量是至关重要的。

\subsubsection{BioBART}

对于生物医学领域，我们采用了BioBART论文中介绍的BART~\cite{DBLP:conf/bionlp/YuanYGZXY22/biobart}结构。
通过领域自适应预训练(Doamin-Adaptive Pre-Training, DAPT)，我们把BART-large应用在PubMed摘要\footnotemark[11]语料上，其中包含大约41GB的生物医学研究论文的摘要。
接着，我们收集各种生物医学语言的生成任务，包括对话, 文本摘要, 实体链接和命名实体识别，用作下游任务表现的评估。
BioBART在不同的基准上都取得了很大的改进，并且与现有的最先进的方法相比，取得了具有竞争力或优越的结果。

\footnotetext[11]{\url{https://pubmed.ncbi.nlm.nih.gov/}; 于 2021 年访问。}

\begin{itemize}
    \item Yuyuan-Bart-139M
    \item Yuyuan-Bart-400M
\end{itemize}

\subsubsection{GPT2}

我们采用与Wenzhong-GPT2-3.5B相同的架构，在50GB的医学(PubMed)语料库上进行预训练。
我们的Yuyuan-GPT2-3.5B是医疗领域最大的开源的GPT2模型。
进一步地，模型可以通过计算困惑度（PPL）来判断事实。
为了完成问答功能，我们将短语模式从疑问的形式转换为了陈述。

\begin{itemize}
    \item Yuyuan-GPT2-3.5B
    \item YuyuanQA-GPT2-3.5B
\end{itemize}

\subsection{系列名待定 (探索)}
\label{sec:tbd_zh}

我们也会与其他组织一同进行一些实验性的探索。
例如，以下模型是我们与追一科技共同努力的成果。

\begin{itemize}
    \item Zhouwenwang-Unified-1.3B
    \item Zhouwenwang-Unified-110M
\end{itemize}

\section{封神框架}
\label{sec:framework_zh}

为了解决第\ref{sec:intro_zh}节中提到的问题，我们结合了Huggingface~\cite{DBLP:journals/corr/abs-1910-03771/huggingface}, Megatron-LM~\cite{DBLP:journals/corr/abs-1909-08053/Megatronlm}, PyTorch-Lightning, 和 DeepSpeed~\cite{DBLP:conf/sc/RajbhandariRRH20/deepspeed}的优势并且融合进了封神框架中。
如果用户熟悉上述框架，那么他们可以毫不费力地使用我们的深度学习框架。
我们的框架支持对TB级数据，以及超过10亿的参数的大规模的模型进行预训练，并且支持对一系列的下游任务进行微调。
通过一些配置，用户可以简单地进行分布式训练和使用节省内存的技术，让用户更加专注于模型部署和创新。
封神框架还提供了我们在Huggingface中的开源的模型及其应用的代码和示例。

我们的框架具有以下优势：
\begin{itemize}
    \item 具有优于原始的Torch库的卓越性能，比如训练性能提升约$300\%$。
    \item 支持大规模的模型：支持百亿级别内模型训练及微调。
    \item 支持超大规模的数据集(TB级)。
    \item 通过提供大量的预训练和下游任务的示例等，使得训练过程易于使用。
    \item 适配各种环境，比如支持在CPU, GPU, TPU等不同设备上运行。
    \item 集成主流的分布式训练逻辑，无需修改代码即可支持DDP和Zero Optimizer等分布式优化技术。
\end{itemize}

\begin{figure}
\small
\begin{forest}
  for tree={
    font=\ttfamily,
    grow'=0,
    child anchor=west,
    parent anchor=south,
    anchor=west,
    calign=first,
    inner xsep=7pt,
    edge path={
      \noexpand\path [draw, \forestoption{edge}]
      (!u.south west) +(7.5pt,0) |- (.child anchor) pic {folder} \forestoption{edge label};
    },
    before typesetting nodes={
      if n=1
        {insert before={[,phantom]}}
        {}
    },
    fit=band,
    before computing xy={l=15pt},
  }  
[fengshen
  [data
    [cbart\_dataloader]
    [fs\_datasetss]
    [universal\_datamodule]
    [megatron\_dataloader]
    [mmap\_dataloader]
    [task\_dataloader]
  ]
  [examples
  ]
  [metric
  ]
  [losses
  ]
  [tokenizer
  ]
  [models
  ]
  [utils
  ]
]
\end{forest}
\caption{封神框架的概览。}
\label{fig:file_tree_zh}
\end{figure}

\begin{figure*}
\centering
\includegraphics[width=\textwidth]{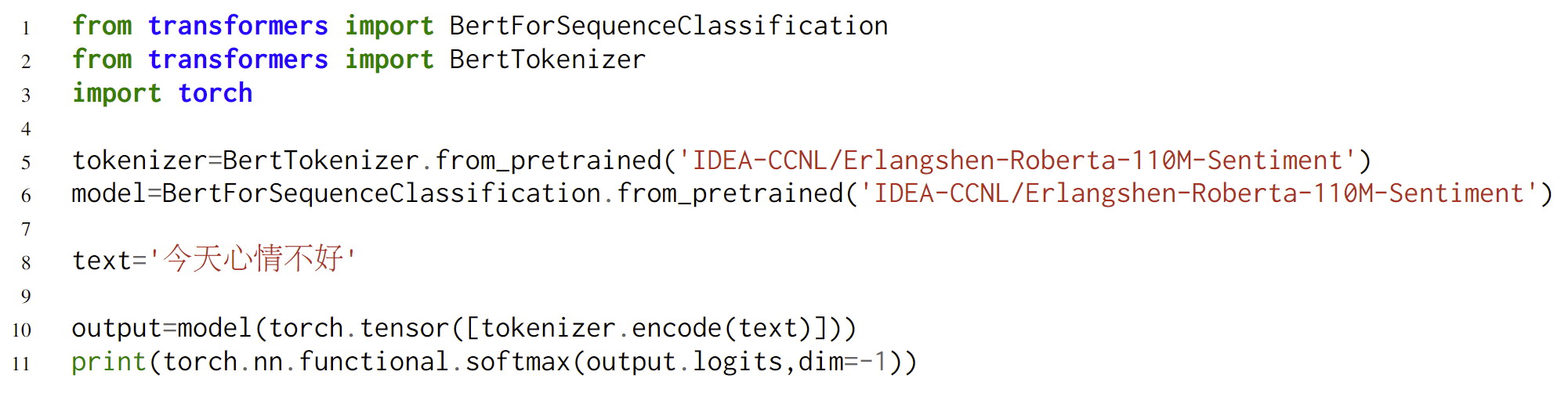}
\caption{使用Erlangshen-Roberta-110M-Sentiment做推理的示例。}
\label{fig:code_example_zh}
\end{figure*}

\subsection{构架}

封神框架中的核心组件如图\ref{fig:file_tree_zh}所示。
我们的模型的核心功能对应图中的各个模块。
如果在封神框架上进行开发，整体上可以按照下面的三个步骤进行：

\begin{enumerate}
    \item 封装数据处理流程
    \item 封装模型结构
    \item 配置一些插件
\end{enumerate}

\subsection{文档}

我们的文档\footnotemark[12]提供了分布式模型训练、微调以及各种大模型应用的使用细节。
此外，我们也展示了参与竞赛的论文以及各种可复现的代码和教程。

\footnotetext[12]{\url{https://fengshenbang-doc.readthedocs.io/zh/latest/index.html}}

\subsection{示例}

如图\ref{fig:code_example_zh}所示，在封神框架中构建模型只需要几行代码。
对于已经在使用Huggingface之类的框架的用户来说，几乎不需要任何成本即可开始使用。

\section{封神榜单}
\label{sec:benchmark_zh}

对于评估模型能力来说，好的基准是必要的。
应该注意的是，中文在语言上与英文和其他印欧语系不同。
因此，我们提出了封神榜单，它是由不同类型的中文自然语言任务组成的中文榜单。
除了在任务中考虑公平的环境之外，我们还计划发布一些通用的预训练数据集。

\subsection{任务选择的标准}

为了构建高质量和健壮的基准，我们需要考虑到如何测试模型的方方面面。 
因此，我们在构建封神榜单时确定了以下要求：

\noindent{\ul 广泛认可。} 
虽然一些现有的数据集不是用中文设计的，但它们多年来在NLP领域中被广泛使用，例如SuperGLUE~\cite{DBLP:conf/nips/WangPNSMHLB19/superglue}。
所以，我们将召集一些专业的中英文的语言专家，精心翻译并校对这些热门的数据集。

\noindent{\ul 面向未来。} 
事实上，一些NLP模型已经在多个基准测试中超越了人类的表现。
这宣告了人工智能已经拥有甚至可以超越人类水平的认知智能。
我们认为的原因之一是这些基准的评估范围有限。
更紧迫和必要的工作是构建具有挑战性的数据集，而不是将现有数据集拟合到 $100\%$ 的准确度。
未来的基准需要考虑更广泛的道德, 技术和社会上的挑战。
我们的数据集将会尽快发布，以更好地支持中文社区的进一步研究。

\noindent{\ul 合作共创。} 
基准需要反映真实世界的场景。
我们希望能够与行业活跃的公司合作收集真实世界的数据并发布。

\subsection{排行榜}

\subsubsection{Chinese-SuperGLUE}

作为一个非常强大的基准，SuperGLUE被整个NLP社区广泛认可，但它仅提供英文版本。
这些高质量的数据集也是中文社区所需要的。
因为如果缺乏对英文数据集的验证，面向中文的模型可能会受到质疑。
这是非常不公平的，甚至可能导致研究仅限于单语环境。
我们认为，一个原因是很难评估不同数据集的质量，导致一些审稿员对一些没见过的数据集感到惊讶。
为了解决上述问题，我们计划开发 Chinese-SuperGLUE 来评估中国模型。
相关论文和榜单即将推出。

\subsubsection{QAKM}

为了评估 NLP 模型的知识水平，我们提出了 QAKM（知识模型问答任务）。
知识模型需要学习特定领域的知识并回答给定数据集中没见过的问题。
该任务已包含在 NLPCC\footnotemark[13] 中，可在此网站\footnotemark[14] 中访问。

\footnotetext[13]{\url{http://tcci.ccf.org.cn/conference/2022/cfpt.php}}
\footnotetext[14]{\url{https://idea.edu.cn/ccnl-act/nlpcc-track1.html}}

\section{总结和未来工作}
\label{sec:future_zh}

本报告介绍我们的开源项目，封神榜，旨在成为中文认知智能的基础设施。
封神榜的三个子项目(封神榜模型、封神框架、封神榜单)支持中文智能系统的各方面进步。
此外，我们要强调的是，我们的封神榜项目是一个持续的开源项目，即我们会不断更新所有子项目。
我们希望个人和组织的贡献者也可以一起加入，共同完善该项目，共同构建整个中文社区。
这个世界需要一些好的想法。

\section*{伦理考量}
\label{sec:ethical_zh}

我们的封神榜项目提供了整个中文生态系统来使用，生产和评估大模型。并且，我们注意到许多公司和研究机构已经部署了我们的模型。
我们的模型和基准正逐渐迁移到现实世界，从而对人类产生无法预计的影响。
考虑伦理影响有很多方面需要关注：大规模模型的隐含偏见, 潜在的环境问题, 标记数据中的不良预判, 开源框架的不当使用等。
为了更好地理解和更深入地讨论伦理问题，我们将在下一版报告中提供详细的讨论。
我们鼓励任何开发者使用封神榜项目中的任何东西时就其使用展开公开辩论，例如任务如何选择和如何部署。
我们希望这将减少任何不当行为发生的机会。

\section*{致谢}
我们感谢CCNL的GTS团队提供的反馈和帮助。

\newpage
\bibliography{anthology,custom}
\bibliographystyle{acl_natbib}

\appendix
\label{sec:append}

\section{A List of Fengshenbang Models}
\label{append:all_model}

This list is in alphabetical order.

\begin{enumerate}
    \item Erlangshen-DeBERTa-v2-186M-Chinese-SentencePiece: \\ \url{https://huggingface.co/IDEA-CCNL/Erlangshen-DeBERTa-v2-186M-Chinese-SentencePiece}
    \item Erlangshen-DeBERTa-v2-320M-Chinese: \\
    \url{https://huggingface.co/IDEA-CCNL/Erlangshen-DeBERTa-v2-320M-Chinese}
    \item Erlangshen-DeBERTa-v2-710M-Chinese: \\ \url{https://huggingface.co/IDEA-CCNL/Erlangshen-DeBERTa-v2-710M-Chinese}
    \item Erlangshen-DeBERTa-v2-97M-CWS-Chinese: \\ \url{https://huggingface.co/IDEA-CCNL/Erlangshen-DeBERTa-v2-97M-CWS-Chinese}
    \item Erlangshen-DeBERTa-v2-97M-Chinese: \\
    \url{https://huggingface.co/IDEA-CCNL/Erlangshen-DeBERTa-v2-97M-Chinese}
    \item Erlangshen-Longformer-110M: \\ \url{https://huggingface.co/IDEA-CCNL/Erlangshen-Longformer-110M}
    \item Erlangshen-Longformer-330M: \\ \url{https://huggingface.co/IDEA-CCNL/Erlangshen-Longformer-330M}
    \item Erlangshen-MegatronBert-1.3B: \\ \url{https://huggingface.co/IDEA-CCNL/Erlangshen-MegatronBert-1.3B}
    \item Erlangshen-MegatronBert-1.3B-NLI: \\ \url{https://huggingface.co/IDEA-CCNL/Erlangshen-MegatronBert-1.3B-NLI}
    \item Erlangshen-MegatronBert-1.3B-Sentiment: \\ \url{https://huggingface.co/IDEA-CCNL/Erlangshen-MegatronBert-1.3B-Sentiment}
    \item Erlangshen-MegatronBert-1.3B-Similarity: \\ \url{https://huggingface.co/IDEA-CCNL/Erlangshen-MegatronBert-1.3B-Similarity}
    \item Erlangshen-MegatronBert-3.9B-Chinese: \\
    \url{https://huggingface.co/IDEA-CCNL/Erlangshen-MegatronBert-3.9B-Chinese}
    \item Erlangshen-Roberta-110M-NLI: \\ \url{https://huggingface.co/IDEA-CCNL/Erlangshen-Roberta-110M-NLI}
    \item Erlangshen-Roberta-110M-Sentiment: \\ \url{https://huggingface.co/IDEA-CCNL/Erlangshen-Roberta-110M-Sentiment}
    \item Erlangshen-Roberta-110M-Similarity: \\ \url{https://huggingface.co/IDEA-CCNL/Erlangshen-Roberta-110M-Similarity}
    \item Erlangshen-Roberta-330M-NLI: \\ \url{https://huggingface.co/IDEA-CCNL/Erlangshen-Roberta-330M-NLI}
    \item Erlangshen-Roberta-330M-Sentiment: \\ \url{https://huggingface.co/IDEA-CCNL/Erlangshen-Roberta-330M-Sentiment}
    \item Erlangshen-Roberta-330M-Similarity: \\ \url{https://huggingface.co/IDEA-CCNL/Erlangshen-Roberta-330M-Similarity}
    \item Erlangshen-Ubert-110M-Chinese: \\ \url{https://huggingface.co/IDEA-CCNL/Erlangshen-Ubert-110M-Chinese}
    \item Erlangshen-Ubert-330M-Chinese: \\ \url{https://huggingface.co/IDEA-CCNL/Erlangshen-Ubert-330M-Chinese}
    \item Erlangshen-ZEN1-224M-Chinese: \\ \url{https://huggingface.co/IDEA-CCNL/Erlangshen-ZEN1-224M-Chinese}
    \item Erlangshen-ZEN2-345M-Chinese: \\ \url{https://huggingface.co/IDEA-CCNL/Erlangshen-ZEN2-345M-Chinese}
    \item Erlangshen-ZEN2-668M-Chinese: \\ \url{https://huggingface.co/IDEA-CCNL/Erlangshen-ZEN2-668M-Chinese}
    \item Randeng-BART-139M: \\ 
    \url{https://huggingface.co/IDEA-CCNL/Randeng-BART-139M}
    \item Randeng-BART-139M-SUMMARY: \\ \url{https://huggingface.co/IDEA-CCNL/Randeng-BART-139M-SUMMARY}
    \item Randeng-BART-759M-Chinese-BertTokenizer: \\
    \url{https://huggingface.co/IDEA-CCNL/Randeng-BART-759M-Chinese-BertTokenizer}
    \item Randeng-MegatronT5-770M: \\ \url{https://huggingface.co/IDEA-CCNL/Randeng-MegatronT5-770M}
    \item Randeng-Pegasus-238M-Chinese: \\ \url{https://huggingface.co/IDEA-CCNL/Randeng-Pegasus-238M-Chinese}
    \item Randeng-Pegasus-238M-Summary-Chinese: \\ \url{https://huggingface.co/IDEA-CCNL/Randeng-Pegasus-238M-Summary-Chinese}
    \item Randeng-Pegasus-523M-Chinese: \\ \url{https://huggingface.co/IDEA-CCNL/Randeng-Pegasus-523M-Chinese}
    \item Randeng-Pegasus-523M-Summary-Chinese: \\ \url{https://huggingface.co/IDEA-CCNL/Randeng-Pegasus-523M-Summary-Chinese}
    \item Randeng-T5-77M: \\ 
    \url{https://huggingface.co/IDEA-CCNL/Randeng-T5-77M}
    \item Randeng-T5-784M: \\ 
    \url{https://huggingface.co/IDEA-CCNL/Randeng-T5-784M}
    \item Randeng-Transformer-1.1B-Denoise: \\
    \url{https://huggingface.co/IDEA-CCNL/Randeng-Transformer-1.1B-Denoise}
    \item Taiyi-CLIP-RoBERTa-326M-ViT-H-Chinese: \\
    \url{https://huggingface.co/IDEA-CCNL/Taiyi-CLIP-RoBERTa-326M-ViT-H-Chinese}
    \item Taiyi-CLIP-Roberta-102M-Chinese: \\ \url{https://huggingface.co/IDEA-CCNL/Taiyi-CLIP-Roberta-102M-Chinese}\
    \item Taiyi-CLIP-Roberta-large-326M-Chinese: \\ \url{https://huggingface.co/IDEA-CCNL/Taiyi-CLIP-Roberta-large-326M-Chinese}
    \item Taiyi-Roberta-124M-D: \\ \url{https://huggingface.co/IDEA-CCNL/Taiyi-Roberta-124M-D}
    \item Taiyi-Roberta-124M-D-v2: \\ \url{https://huggingface.co/IDEA-CCNL/Taiyi-Roberta-124M-D-v2}
    \item Taiyi-vit-87M-D: \\ \url{https://huggingface.co/IDEA-CCNL/Taiyi-vit-87M-D}
    \item Wenzhong-GPT2-110M: \\ \url{https://huggingface.co/IDEA-CCNL/Wenzhong-GPT2-110M}
    \item Wenzhong-GPT2-3.5B: \\ \url{https://huggingface.co/IDEA-CCNL/Wenzhong-GPT2-3.5B}
    \item Wenzhong2.0-GPT2-3.5B-chinese: \\ \url{https://huggingface.co/IDEA-CCNL/Wenzhong2.0-GPT2-3.5B-chinese}
    \item Yuyuan-Bart-139M: \\ \url{https://huggingface.co/IDEA-CCNL/Yuyuan-Bart-139M}
    \item Yuyuan-Bart-400M: \\ \url{https://huggingface.co/IDEA-CCNL/Yuyuan-Bart-400M}
    \item Yuyuan-GPT2-3.5B: \\ \url{https://huggingface.co/IDEA-CCNL/Yuyuan-GPT2-3.5B}
    \item YuyuanQA-GPT2-3.5B: \\ \url{https://huggingface.co/IDEA-CCNL/YuyuanQA-GPT2-3.5B}
    \item Zhouwenwang-Unified-1.3B: \\ \url{https://huggingface.co/IDEA-CCNL/Zhouwenwang-Unified-1.3B}
    \item Zhouwenwang-Unified-110M: \\ \url{https://huggingface.co/IDEA-CCNL/Zhouwenwang-Unified-110M}
\end{enumerate}

\section{Author Contributions}
\label{append:author_contribution}

Fengshenbang project is an ongoing open-source effort maintained by the team of engineers, researchers, interns at the Cognitive Computing and Natural Language (CCNL) Research Center in International Digital Economy Academy (IDEA)\footnotemark[15]. 
The project managers initiated the Fengshenbang project with its sub-projects, which includes:
\begin{itemize}
    \item Jiaxing Zhang \\(\href{mailto:zhangjiaxing@idea.edu.cn}{zhangjiaxing@idea.edu.cn})
    \item Ruyi Gan \\(\href{mailto:ganruyi@idea.edu.cn}{ganruyi@idea.edu.cn})
\end{itemize}

\footnotetext[15]{\url{https://www.idea.edu.cn/}}

Other authors have contributed to the project in various ways, including data collection, designing and running experiments, constructing models, etc. 
Therefore, we provide a brief description:
\begin{itemize}
    \item Fengshenbang Model: Ping Yang, Xinyu Gao, Ziwei Wu, Xiaoqun Dong, Junyu Lu, Weifeng Chen, Junjie Wang, Junqing He, Yongfeng Huang, Xiayu Li, Yanghan Wu, Qi Yang, Yuxiang Zhang, Jianheng Zhuo, Xinyu Zhu, Zhongshen Zeng, Ting Han, Rui Wang, Xiaojun Wu, Kunhao Pan, Hao Wang, Chongpei Chen
    \item Fengshen Framework: Xinyu Gao, Ping Yang, Ziwei Wu
    \item Fengshenbang Benchmark: Junjie Wang, Ziwei Wu, Yuxiang Zhang
    \item Paper writing: Junjie Wang \\ (\href{mailto:wjj1020181822@toki.waseda.jp}{wjj1020181822@toki.waseda.jp}), \\ Yuxiang Zhang, Lin Zhang\footnotemark[16]
\end{itemize}

\footnotetext[16]{From CTO Lab in IDEA}

\end{CJK*}
\end{document}